\documentclass{article}
\usepackage{iclr2027_conference,times}

\newif\ifdraft
\draftfalse
\usepackage{microtype}
\usepackage{graphicx}
\usepackage{xcolor}
\usepackage{subcaption} 
\usepackage{booktabs}
\usepackage[hyphens]{url}
\usepackage{tabularx}
\usepackage{multirow}   

\usepackage{enumitem}
\usepackage[linesnumbered,ruled,vlined]{algorithm2e}
\usepackage{amssymb}
\usepackage{amsthm}

\usepackage{amsmath}
\usepackage{threeparttable}
\usepackage{xspace}

\usepackage{tcolorbox}

\newtcolorbox[auto counter]{observation}[1][]{
  colback=black!5,
  colframe=black!25,
  boxrule=0.4pt,
  arc=0pt,
  boxsep=0pt,
  left=8pt,
  right=8pt,
  top=6pt,
  bottom=6pt,
  before skip=8pt,
  after skip=8pt,
  fontupper=\itshape,
  before upper={\textbf{Observation~\thetcbcounter.}\ },
  #1
}

\definecolor{pqred}{RGB}{255,0,102}

\newcommand{\topk}{{LiteTopK}\xspace}

\newcommand{\dsa}{{LiteDSA}}

\usepackage{hyperref}

\title{LiteTopK: Exploiting the Curse of Dimensionality for a Fused Indexer-TopK Kernel in Long-Context Sparse Attention}

\author{
Ziqi Yin \\
Nanyang Technological University \\
\And
Jianyang Gao \\
ETH Zurich \\
\And
Peiqi Yin \\
The Chinese University of Hong Kong \\
\And
Jiangneng Li \\
Nanyang Technological University \\
\And
Gao Cong \\
Nanyang Technological University
}

\ifdraft
\else
  \iclrfinalcopy
  \usepackage{etoolbox}
  \makeatletter
  \patchcmd{\@maketitle}
    {\lhead{Published as a conference paper at ICLR 2027}}
    {\lhead{Preprint}}
    {}{}
  \makeatother
\fi

\begin{document}
\maketitle

\begin{abstract}
Indexer-TopK, which computes scores and selects the top-$k$ candidates, is a key
primitive in long-context sparse attention for large language models, vector
retrieval for recommendation systems, and vector databases. However, existing
GPU-based Indexer-TopK kernels like DeepSeek Sparse Attention (DSA) typically
separate scoring from selection and materialize the full score matrix in high-bandwidth memory. This strategy is inefficient for large score matrices, such as those encountered in long-context sparse attention, due to excessive global-memory traffic, costly synchronization, and prohibitive memory overhead. 

In this study, motivated by the curse of dimensionality, we observe that sparse attention scores exhibit a score concentration phenomenon, with scores falling within a narrow range. Based on this observation, we propose \topk{}, a sample-filter-select framework that estimates query–data score ranges from a small subset of the data. It uses these estimates to filter scores as they are computed, materializes only the candidates that pass the filter, and selects from the surviving candidates. In particular, we develop two implementations of \topk{}. The first uses a conservative threshold estimate, while the second pairs a predicted threshold with a fallback strategy. Building on \topk{}, we further propose \dsa{}, which reuses overlapping key/value entries across neighboring queries to accelerate attention computation. We integrate \topk{} into vLLM, SGLang, and Megatron-LM. Our evaluation on eight B200 GPUs demonstrates prefill speedups of up to 1.49$\times$ across GLM-5.2, DeepSeek-V4-Flash, and Hy-4-Preview at a context length of 1M tokens, without compromising model capabilities. 
\end{abstract}

\section{Introduction}
\label{sec:intro} 

\begin{figure*}[!t]
    \centering
        
    \begin{minipage}[t]{0.49\textwidth}
        
        \centering
        \includegraphics[height=0.49\linewidth]{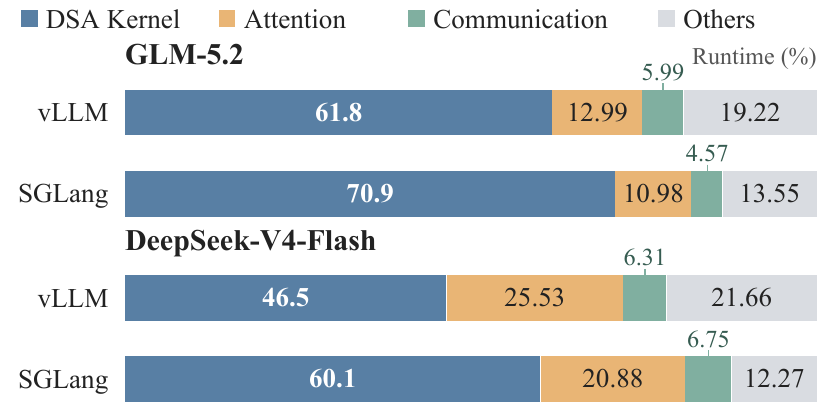}
        \caption{Prefill runtime breakdown at a 1M-token context length for GLM-5.2 and DeepSeek-V4-Flash on eight NVIDIA B200 GPUs using tensor and expert parallelism.}
        \label{fig:breakdown}
    \end{minipage}
    \hfill
    \begin{minipage}[t]{0.49\textwidth}
        
        \centering
        \includegraphics[height=0.49\linewidth]{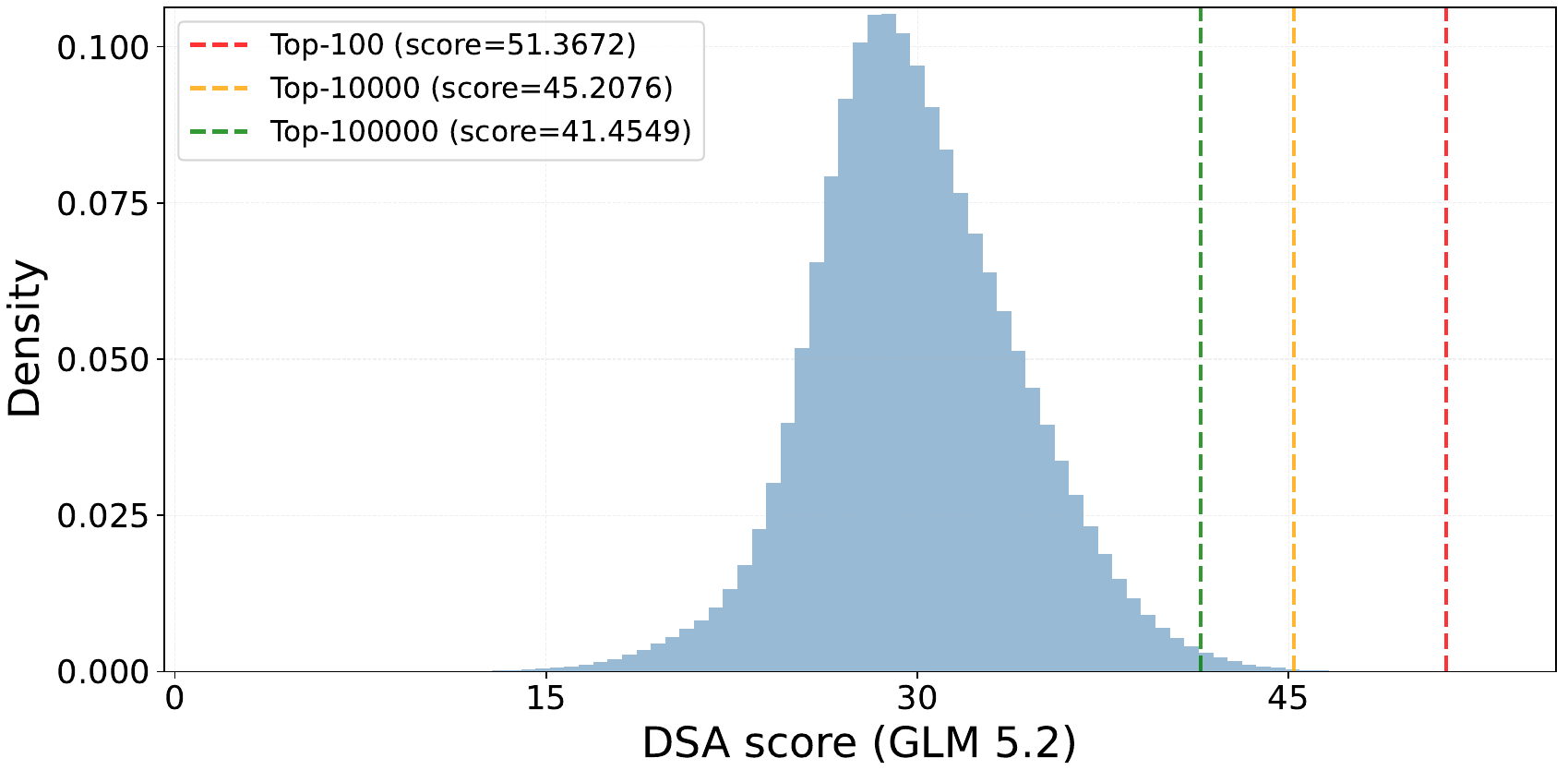}
        \caption{DSA score distribution in the first layer of GLM-5.2 during prefill on the Wiki dataset.}
        \label{fig:score_distribution_dsa}
    \end{minipage}
    
\end{figure*}

Indexer-TopK, which computes scores under a specific scoring function and selects the per-row top-$k$ elements, has become a core primitive in modern machine learning systems, with applications in sparse attention of large language model (LLM)~\citep{tang2024quest,liu2025deepseek,synk2025exploiting}, recommendation systems~\citep{DBLP:conf/www/KhandagaleJASYW25,gao2021learning}, and information retrieval~\citep{lee2023rethinking}. 
Representative examples include: 

\begin{enumerate}[leftmargin=*,topsep=0pt,itemsep=0pt]
\item 
As context lengths grow to millions of tokens, attention computation becomes prohibitively expensive, motivating sparse-attention systems~\citep{SparQ24,chen2024arkvale,liu2025deepseek,synk2025exploiting,tang2024quest} such as DeepSeek Sparse Attention (DSA)~\citep{deepseekai2026deepseekv4} to approximate full attention using only a small subset of tokens. Specifically, 
DSA treats the current text chunk as queries and the historical context as candidate keys and values. It computes and stores lightweight relevance scores between the preceding context and the chunk text, then applies top-$k$ selection to identify the most relevant historical tokens for constructing sparse attention. While this design reduces the cost of full attention, its score-computation and selection pipeline itself becomes a major performance bottleneck. As shown in Figure~\ref{fig:breakdown}, the DSA kernel accounts for 70.9\% of the total prefill runtime of GLM-5.2~\citep{glm5team2026glm5vibecodingagentic} in SGLang~\cite{zheng2024sglang} at a 1M-token context length.

\item Recommendation systems and information retrieval systems typically encode queries/users and candidate items into high-dimensional embeddings using deep learning models~\citep{DBLP:conf/www/KhandagaleJASYW25,gao2021learning}. During candidate retrieval, they use Indexer-TopK to identify the top-$k$ candidates with the highest similarity scores, where $k$ can be on the order of 
thousands. The retrieved candidates are then re-ranked by more sophisticated models. 
\end{enumerate}

Despite the broad applications, existing Indexer-TopK implementations remain inefficient on GPUs because they typically decouple score computation and selection into two stages: first, computing and materializing the full score matrix in high-bandwidth memory (HBM) and then performing a top-$k$ selection, such as radix-select~\citep{alabi2012fast}. This strategy introduces two major overheads. First, it incurs substantial {memory overhead}. For example, the DSA kernel incurs a runtime memory overhead of 32 GB 
under a 1M-token prefill with a moderate chunk size (e.g., 8192 in vLLM~\cite{kwon2023efficient}). This is because the kernel needs to keep the full score matrix before the top-$k$ selection. 
Such runtime memory overhead reduces the portion of HBM for storing the KV cache, which is already tight in existing LLM serving systems. 
To mitigate this, 
vLLM and SGLang 
further split the 8192-token chunk into smaller sub-chunks during prefill, such as 512, 
to reduce the HBM footprint, 
but at the cost of lower prefill throughput. 
Second, it incurs significant {latency overhead}, 
since the score matrix must be written back to HBM and subsequently read multiple times during Top-$k$ selection, leading to significant latency. 
The same issue is also noted in the vLLM blog\footnote{\url{https://vllm.ai/blog/2025-09-29-deepseek-v3-2}}, 
which observes that ``a clear challenge is that at high batch size with long context, the logits tensor is materialized before running a row-wise top-$k$.''




To address this, we 
study the distribution of the scores in sparse attention and observe that these scores consistently show a concentration phenomenon, which is brought by the curse of dimensionality in high-dimensional spaces~\citep{curse1,curse2}. In particular, similarity scores between high-dimensional vectors, including Euclidean distance and inner product, tend to concentrate within a narrow range while exhibiting an extremely long-tailed distribution, which is widely observed in real-world datasets~\citep{ bruch2024foundations}. Since sparse attention scores are computed 
from the similarity scores between high-dimensional vectors, this concentration effect also appears in sparse attention. For example, DSA applies ReLU-based weighting to inner-product scores, resulting in a distribution that exhibits the same narrow-range concentration and long-tail behavior. As shown in Figure~\ref{fig:score_distribution_dsa}, DSA scores in the first layer of GLM-5.2 under a 1M-token context are concentrated between 25 and 35, whereas scores of 45 or higher are extremely rare. Similar also patterns are observed across different models and layers, as shown in Appendix~\ref{sec:pattern_appendix}.


Building on this pattern, we propose \topk{}, a sample-filter-select framework that fuses candidate filtering into indexer scoring. It estimates query–data score ranges from a small subset of the data and uses these estimates to filter candidates as their scores are computed. Only candidates that pass the filter are materialized for final selection. In particular, we propose two different  implementations within this framework. The first implementation samples a small candidate subset to estimate each query token's score distribution and constructs an equal-width histogram. The bin containing the local $k$-th largest score defines the initial threshold bin. During filtering, \topk{} computes scores for all candidates but materializes only those whose scores meet or exceed the threshold bin's lower bound, while idle warps lazily update the threshold. During selection, candidates in higher-scoring bins are output directly, and further selection is restricted to the threshold bin. The conservatively chosen initial threshold bin preserves correctness without requiring a fallback, but may admit more candidates than necessary, increasing candidate maintenance and selection costs. The second implementation replaces the conservative thresholding strategy with a sampled-quantile predictor to estimate the global top-$k$ threshold. This yields a more aggressive threshold that can reduce the number of retained candidates and the associated maintenance and selection costs. However, prediction errors may cause candidate-buffer underflow, requiring a fallback strategy to preserve correctness.

Beyond Indexer-TopK, the attention stage often consists of many small, memory-bound GEMMs. We observe that neighboring query tokens frequently select substantially overlapping candidate sets. Exploiting this locality, we propose \dsa{}, which merges neighboring queries' selected key/value entries into a shared, deduplicated list for downstream sparse attention. The grouped queries reuse key/value loads, while per-query membership masks restrict attention to each query's original top-$k$ set. By reducing redundant memory accesses, \dsa{} significantly accelerates attention computation.

In summary, this work makes the following contributions:
\begin{itemize}[leftmargin=*,topsep=2pt,itemsep=2pt]
    \item To the best of our knowledge, we are the first to observe and characterize the score concentration phenomenon in sparse attention, trace its origin to the well-known curse of dimensionality in high-dimensional vector spaces, and further exploit this property to improve sparse attention kernel design. 

     \item We propose \topk{}, which, to the best of our knowledge, is the first sparse-attention Indexer-TopK design to fuse online candidate filtering with indexer scoring without materializing the full score matrix in HBM. Compared with the existing DSA kernel, \topk{} delivers kernel-level speedups of up to $1.72\times$ during prefill and $1.88\times$ during decoding. Its design principles also extend to other Indexer-TopK workloads.

    \item We propose \dsa{}, a sparse-attention kernel that reuses key/value loads across neighboring queries while preserving their individual top-$k$ candidate sets through membership masks. On the GLM-5.2 attention benchmark, \dsa{} achieves up to $1.81\times$ speedup over the origin vLLM attention kernel.

    \item In a real-world deployment on 8 B200 GPUs, \topk{} achieves up to a 1.49$\times$ end-to-end speedup in the prefill stage across GLM-5.2, DeepSeek-V4-Flash, and Hy-4-Preview, while using less memory. The source code of our implementation is available at \url{https://github.com/sgl-project/sglang/pull/47469} and \url{https://github.com/vllm-project/vllm/pull/65920}.
\end{itemize}

\section{BACKGROUND}
\label{sec:background}

\noindent\textbf{Sparse Attention.} As million-token context windows are increasingly common in real-world LLM production environments, sparse attention algorithms~\citep{lai2026minimax,yuan2025native,liu2025deepseek,deepseekai2026deepseekv4,tang2024quest,SparQ24, glm5team2026glm5vibecodingagentic,tencenthyteam2026hy4preview} has become indispensable for reducing the I/O and computational costs of full attention. A representative example is DeepSeek Sparse Attention (DSA)~\citep{liu2025deepseek}, a natively trained sparse attention module, which has become the foundation for massively deployed models such as DeepSeek-V3.2/V4~\cite{liu2025deepseek,deepseekai2026deepseekv4}, GLM-5 family~\citep{glm5team2026glm5vibecodingagentic}, Hy-4-Preview~\cite{tencenthyteam2026hy4preview}, Qwen3.8-Flash-Next~\cite{qwen3.8flashnext}, and Longcat-2.0~\cite{longcat2026longcat2}. 
DSA follows a score-then-select paradigm. It develops a lightweight indexer, which stores an indexer key vector $k_s \in \mathbb{R}^d$ 
for every preceding token $s$. For a query token $t$, it derives $H$  indexer heads $q_{t,j}\in \mathbb{R}^d,j=1,...,H$ and aggregates their head-wise scores to compute the index score as follows:
\begin{equation}
I_{t,s}
=
\sum_{j}^{H} w_{t,j}\,
\mathrm{ReLU}\!\left(
(q_{t,j})^{\top} k_{s}
\right).
\label{eq:dsa_score}
\end{equation}
where $w_{t,j}$ is a query-dependent weight used to aggregate the per-head scores. A per-query top-$k$ selection with $k=2,048$ is then applied to the aggregated score $I_{t,s}$, and full attention is computed over the Key/Value vectors of the selected top-2,048 tokens.

Another class of sparse-attention methods is training-free, meaning that they can be directly applied to off-the-shelf LLM checkpoints without modifying model weights or requiring sparse-attention pretraining~\cite{yang2025less,TidalDecode,li2024snapkv}. These methods typically follow the 
same design: performing score-then-select independently for each attention head. They mainly differ in how per-head relevance scores are computed, ranging from exact query-key inner products~\citep{gupta2021memory,TidalDecode} to approximate scores derived from reduced representations, such as salient query channels~\citep{SparQ24} and product quantization~\citep{zhang2025pqcache}. However, these training-free methods typically introduce non-negligible accuracy degradation~\cite{nawrot2026sparse,yuan2025native}. In this study, we integrate \topk with DSA, 
while \topk is broadly applicable to other score-then-select sparse-attention paradigms as well. It is also worth noting that \topk is compatible with both training and inference systems (See Section 3.2).

\noindent\textbf{LLM System.} To support the training and inference of large language models, LLM systems have become essential infrastructure, with representative systems including vLLM~\cite{kwon2023efficient}, SGLang~\cite{zheng2024sglang}, Megatron-LM~\cite{megatron-lm}, and FlashInfer~\cite{ye2025flashinfer}. These systems aim to improve efficiency, including throughput and latency, as well as GPU memory utilization, while preserving algorithmic correctness. Existing efforts can be broadly categorized along four main dimensions: (1) reducing GPU memory traffic, as in FlashAttention~\cite{dao2022flashattention}, which tiles attention and applies online softmax so that the attention score matrix is never materialized in HBM; (2) increasing parallelism, as in FlashDecoding~\cite{hong2024flashdecoding}, which splits the KV dimension across CTAs to better saturate GPU SMs when the number of queries is small; (3) overlapping computation with data movement, as in FlashAttention-3~\cite{shah2024flashattention}, which uses warp-specialized pipelines and asynchronous copies to hide memory-access latency; and (4) reducing redundant memory overhead, as in PagedAttention~\cite{kwon2023efficient} and RadixAttention~\cite{zheng2024sglang}, which respectively manage KV cache with paged memory allocation and reuse shared prefix states across requests. Despite the importance of GPU memory management, to the best of our knowledge, no prior LLM serving system has explored fusing indexer scoring with top-$k$ selection to avoid materializing intermediate scores in HBM. To fill this gap, we propose LiteTopK, which not only reduces the latency but also significantly reduces memory overhead. 


\noindent\textbf{GPU Architecture.} Parallelism is fundamental to modern high-performance GPU computation. One source of such parallelism comes from the heterogeneous computational units: modern NVIDIA GPUs use Tensor Cores for matrix-matrix multiplication and CUDA cores for scalar and element-wise operations. Since these units have separate pipelines and can execute concurrently, per-score scalar operations can overlap with Tensor Core score computation. For example, in DSA, the dot products are computed on Tensor Cores, while the ReLU and weighted aggregation are performed on CUDA cores. Another source is thread-level parallelism: GPU kernels are organized into Cooperative Thread Arrays (CTAs), each containing multiple warps. Beginning with Hopper, warps within a CTA can assume specialized roles: producer warps load tiles from HBM into shared memory, while consumer warps process previously loaded tiles to compute scores. This forms a software pipeline that overlaps data movement with computation. \topk{} exploits these pipeline gaps to hide its additional operations.






\section{Methodology}
\label{sec:method}

\subsection{Problem Definition}
\label{sec:overview}

\noindent\textbf{Problem Definition.} Given a query set $Q$ and a candidate set $X$, our objective is to identify, for each query $q$, the $k$ candidates with the highest scores under a scoring function $f$:
\begin{equation}
\mathrm{TopK}(q) = \operatorname*{arg\,topk}_{x \in X} f(q, x), \qquad q \in Q.
\label{eq:topk}
\end{equation}
This paper focuses on sparse attention, where $f$ is an indexer score, such as  the DSA score defined in Equation~\ref{eq:dsa_score}. We also consider other settings, including $k$ nearest neighbor (KNN) search. We next describe \topk{} and \dsa{} in detail.

\subsection{LiteTopK}
\label{sec:litetopk}

\noindent\textbf{Sampling.} We sample individual keys at regular intervals, leveraging native TMA support for strided loads on Blackwell GPUs. This is because high-scoring keys often occur in contiguous runs, so block-based sampling can overrepresent or miss these runs, leading to noisy threshold estimates. Our earlier evaluation showed that selecting the most frequently occurring hot samples from the previous prefill chunk's top-k yields a tighter threshold, reducing subsequent maintenance and selection overhead. However, we later found that random sampling delivers greater overall performance gains. Although it requires more samples to achieve the same result, its fixed sampling positions allow a persistent kernel to preload the KV cache needed by the next kernel, reducing pipeline bubbles.

After obtaining the sample set, we score the sampled candidates against each query. Here, we propose to subtract a query-specific reference score $s_r$  from each score $s$, yielding the centered score $s-s_r$. This preprocessing step reduces the magnitude of scores dominated by a large common offset while preserving their ordering, helping mitigate quantization loss in low-precision representations. We add the reference back when reconstructing the filtering threshold.

The two implementations differ in how they construct the filtering threshold. The first determines the minimum and maximum sample scores, $s_{\min}$ and $s_{\max}$, then partition
$[s_{\min},s_{\max}]$ into $m$ equal-width bins, with 
$\Delta=(s_{\max}-s_{\min})/m$ and $\delta=1/\Delta$. A score $s$ is
assigned to bin
\[
b(s)=\min\!\left\{m-1,\,
\max\!\left\{0,\left\lfloor(s-s_{\min})\delta\right\rfloor\right\}
\right\},
\qquad \delta=\frac{1}{\Delta}.
\]. 

We initialize the histogram with the sample's bin counts and set the threshold bin to the bin containing
the $k$-th largest sampled score. During filtering, a candidate
passes the gate if its bin ID is at least the threshold bin ID, which is conservative and preserves correctness.




In the second implementation, we estimate the global top-$k$ cutoff using a sample quantile. For a query with $N$ visible keys and $S$ valid samples, we define
\[
j = \frac{kS}{N},
\qquad
j' = \left\lceil \left(1+4\sqrt{\frac{2}{j}}\right)j \right\rceil
   = \left\lceil j+4\sqrt{2j} \right\rceil.
\]
Here, we use the $j'$-th largest sampled score as the predicted filtering threshold. The statistical rationale is as follows. Under an independent uniform-sampling approximation, the number of sampled keys belonging to the global top-$k$ has mean $j$ and standard deviation at most $\sqrt{j}$. This motivates adding a safety margin proportional to $\sqrt{j}$ to the nominal rank, with an empirically chosen coefficient.\footnote{In this line of implementation, sampled-quantile threshold prediction was first introduced by Roberto L. Castro in vLLM PR~\#60618 on October 8, 2026, using an inflated sample rank (e.g., $\lceil1.5j\rceil$ in one configuration) to construct the predicted filtering threshold. Ziyi Xu subsequently refined the rank-selection rule in SGLang PR~\#47469, using the inflated sample rank $j'$ to determine the filtering threshold.}

\noindent\textbf{Filtering.} We fuse filtering into the scoring pipeline: as soon as a score is computed, we compare it with the current threshold. In particular, for the first implementation, we map it to a floating-point bin coordinate $B_{t,s}$, and retain the candidate only if its
bin ID meets the current threshold. The implementation focuses on two operations: computing bin IDs and writing surviving candidates to the shared-memory buffer. For DSA, we incorporate the affine bin mapping directly into the existing weighted sum:
\begin{equation}
\begin{aligned}
B_{t,s}
&= (I_{t,s}-s_{\min})\delta \\
&= \sum_j w'_{t,j}\,
   \mathrm{ReLU}\!\left(q_{t,j}^{\top}k_s\right)
   -s_{\min}\delta,
\qquad
w'_{t,j}=w_{t,j}\delta.
\end{aligned}
\label{eq:bin}
\end{equation}
Since $s_{\min}$ and $\delta$ remain fixed throughout the scan
for each query, we pre-scale the weights once before loading
them into registers. We then initialize the weighted-sum
accumulator to $-s_{\min}\delta$ instead of zero. The existing
FFMA chain thus produces $B_{t,s}$ directly, without a separate
affine transform for each score. We convert this value to a
bin ID and retain the candidate if its bin ID is at least the
current threshold bin ID. Storing the floating-point value $B_{t,s}$ preserves the
within-bin score information needed for final ranking. Because
$\delta>0$, ranking by $B_{t,s}$ is equivalent to ranking by
$I_{t,s}$ in exact arithmetic. At the final output stage,
scores are converted back to their original scale using
$I_{t,s}=B_{t,s}/\delta+s_{\min}$. In the second implementation, filtering requires only a floating-point comparison.

\noindent\textbf{Threshold Refresh.} This  stage is required only in the first implementation. Each passing candidate  atomically increments its bin in the per-query shared-memory histogram. The return values of these increments are unused, avoiding a direct data dependency in subsequent score computations. As the histogram grows, the gate can tighten beyond its initial setting. Periodically, an idle warp, such as warp 10 or 11 in the DSA producer-consumer implementation on Blackwell GPUs, recomputes the threshold bin
by accumulating histogram counts from the highest bin downward until the
cumulative count reaches $k$. The new value is published with a single plain store to one shared-memory slot, with no atomic operation and no barrier. Computing warps independently refresh their cached threshold at fixed intervals and continue using the previous value between reads, so the update does not disrupt computation execution. A stale threshold only loosens the gate, rarely admitting extra candidates under score concentration, but does not affect accuracy. Threshold maintenance therefore remains off the critical path, and does not stall the pipeline. In the second implementation, the predicted threshold remains fixed throughout the scan, so filtering requires only a floating-point comparison.

\noindent\textbf{Candidate Maintenance.} In the first implementation, each lane within each warp maintains a small unordered staging list in shared memory. Candidates that pass the gate are appended to lane-private slots without contending for a shared write pointer. Periodically, the warp writes its staged candidates to the global buffer in batches. For each query, one atomic update reserves space for the whole batch, and each lane writes its candidates to its assigned positions. If a lane's staging list fills up before the next flush, it writes additional candidates directly to the global buffer.

In the second implementation, the tighter predicted threshold typically produces a sparser candidate set. We exploit this sparsity through warp-level voting: if no lane has a passing candidate, the warp skips candidate staging for that batch. Otherwise, it records the passing lanes in ballot masks and stores their scores in fixed shared-memory slots. Dedicated drain warps, corresponding to the idle warps used for threshold refresh in the first implementation, periodically use these masks to compact the surviving candidates. For each query's nonempty batch, a single atomic counter update reserves a contiguous region in the global candidate buffer for the scores and corresponding key indices. 

\noindent\textbf{Top-$k$ Selection.}
Once scoring is complete, \topk{} selects the top-$k$ entries from the surviving candidates stored in the candidate buffer.


\noindent\textbf{Extension to Decoding.} Decoding has a different cost profile from prefill. Prefill processes large token chunks and therefore incurs the costs of both (1) writing a large score matrix back to HBM and (2) performing selection over that matrix. By contrast, decoding generates only one token per step, largely eliminating the write-back cost and leaving top-$k$ selection as the bottleneck. We therefore fuse selection directly into score computation. Each CTA maintains a local histogram whose bins are derived from the intrinsic bit representation of the scores rather than from sampled value ranges, and each computed score triggers a fire-and-forget atomic increment to the corresponding bin. After all scores have been computed, we aggregate the per-CTA histograms to identify the threshold bin and then restrict the final selection to elements within that bin.

\subsection{LiteDSA}
\label{sec:litedsa}

The key idea of \dsa{} is to avoid letting each token scan its selected key/value vectors independently. Motivated by Observation~\ref{obs:cross-chunk-locality}, we group 16 adjacent query tokens and merge their selected candidates into a shared scan list. The group reuses the same key/value loading and computation pipeline, reducing redundant memory traffic. Because the union may contain positions selected by only some queries, we retain a separate membership mask for each query.
\begin{observation}[label=obs:cross-chunk-locality]
Adjacent query tokens tend to select substantially overlapping
sets of prefix tokens in their top-$k$ attention results.
\end{observation}

\noindent\textbf{Candidate Union and Membership Masks.}
For each query group, a CTA first clears a bit-packed bitmap over the candidate position space. During the \topk{} selection phase, the lane handling each selected candidate marks its position using a shared-memory \texttt{atomicOr}. Candidates selected by multiple queries are thereby deduplicated. We then sweep and compact the bitmap into one explicit physical key/value list shared by the group. In a second pass, \dsa{} builds a query-major membership mask over the compacted union slots, indicating which entries belong to each query's original top-$k$ set.

\noindent\textbf{Grouped Sparse Attention.}
All queries in the group compute attention over the shared list. For each query, \dsa{} ANDs the valid-lane mask with that query's membership mask and sets non-member scores to $-\infty$ before softmax. Thus, each query still attends to exactly its own selected candidates: the union enables shared key/value loads, while the masks preserve the original sparse-attention semantics. The benefit depends on candidate overlap; additional union entries incur computation, so \dsa{} targets the neighboring-query locality observed in long-context prefill.

\noindent\textbf{Conservative versus predictive thresholding.} 
Although the conservative implementation avoids fallback recovery, the predictive implementation also rarely triggers recovery in our experiments. Its tighter predicted threshold retains fewer candidates, reducing candidate maintenance and selection costs and improving performance in practice. Predictive thresholding is therefore used in the merging vLLM implementation (\href{https://github.com/vllm-project/vllm/pull/65920}{PR~\#65920})
and SGLang implementation
(\href{https://github.com/sgl-project/sglang/pull/47469}{PR~\#47469}). Below, we report the results of the first implementation. The results of the second implementation are available in the PR referenced above.

\section{Experiments}

\noindent We organize our evaluation around four research questions:
(1) does \topk{} accelerate long-context prefilling end-to-end across vLLM,
SGLang, and Megatron-LM? 
(2) does \topk{} preserve exact Top-$k$ outputs and the long-context capability
of the underlying LLM? 
(3) are these end-to-end prefill speedups driven by lower kernel latency and reduced memory overhead?
(4) do the benefits of \topk{} generalize to other GPU architectures and
workloads, such as decode and vector retrieval?
We additionally evaluate \dsa{} at the attention-kernel level.

\subsection{Experimental Setup}

\label{sec:experimental-setting}


\begin{table*}[t]
\centering
\small

\setlength{\tabcolsep}{3pt}
\caption{End-to-end prefill latency on eight NVIDIA B200 GPUs with TP and EP. Latencies are reported in seconds. Speedup is calculated as Native latency divided by \topk latency. Megatron-LM results for Hy-4-Preview are unavailable because the framework did not yet support this newly released model at the time of evaluation.}

\label{tab:e2e_frameworks}
\renewcommand{\arraystretch}{1.2}
\resizebox{\textwidth}{!}{
\begin{tabular}{
ll
rrr@{\hspace{5pt}}
rrr@{\hspace{5pt}}
rrr
}
\toprule
\multirow{2}{*}{Model}
& \multirow{2}{*}{Context}
& \multicolumn{3}{c}{vLLM}
& \multicolumn{3}{c}{SGLang}
& \multicolumn{3}{c}{Megatron-LM} \\
\cmidrule(lr){3-5}
\cmidrule(lr){6-8}
\cmidrule(lr){9-11}
& & Native & \topk & Speedup
& Native & \topk & Speedup
& Native & \topk & Speedup \\
\midrule
\multirow{4}{*}{GLM-5.2}
& 256K & 18.61 & 17.57 & 1.06$\times$& 20.18& 17.93& 1.13$\times$&17.56 &17.13 & 1.03$\times$ \\
& 512K & 48.00& 42.59& 1.13$\times$&51.94 & 40.67& 1.28$\times$& 45.53& 41.85& 1.09$\times$\\
& 768K & 90.51 & 74.95& 1.21$\times$& 94.03& 72.39& 1.30$\times$& 82.66& 73.86& 1.12$\times$\\
& 1M   & 145.81& 113.27& 1.29$\times$& 148.87& 108.56& 1.37$\times$& 136.29& 113.18& 1.20$\times$\\
\midrule
\multirow{4}{*}{DeepSeek-V4-Flash}
& 256K & 6.92& 6.63& 1.04$\times$& 7.53& 7.08&1.07$\times$ & 8.34& 8.11& 1.03$\times$\\
& 512K & 18.02& 17.18& 1.05$\times$& 21.60& 18.27& 1.18$\times$& 20.90& 19.80& 1.06$\times$\\
& 768K & 33.49	& 31.13& 1.08$\times$& 44.36& 33.63& 1.32$\times$& 40.37& 35.14& 1.15$\times$\\
& 1M   & 53.12& 48.50& 1.10$\times$& 73.06& 53.12& 1.38$\times$&63.89 & 54.75& 1.17$\times$\\
\midrule
\multirow{4}{*}{Hy-4-Preview}
& 256K & 24.38& 23.50 & 1.04$\times$& 23.57 & 22.76& 1.04$\times$& {N/A} & {N/A} & {N/A} \\
& 512K & 59.37& 54.33& 1.09$\times$& 58.71&51.29 & 1.14$\times$& {N/A} & {N/A} & {N/A} \\
& 768K & 106.11& 92.75& 1.14$\times$& 105.61& 86.94& 1.21$\times$& {N/A} & {N/A} & {N/A} \\
& 1M   & 164.91& 138.63& 1.19$\times$&164.92 & 130.29& 1.27$\times$& {N/A} & {N/A} & {N/A} \\
\bottomrule
\end{tabular}
}

\end{table*}

\begin{table*}[t]
\centering
\small
\setlength{\tabcolsep}{3pt}
\caption{
End-to-end prefill latency in SGLang on eight NVIDIA B200 GPUs
with TP=8 and CP=8.
Latencies are reported in seconds.
Speedup = Native / \topk.
}

\label{tab:e2e_sglang_cp}
\renewcommand{\arraystretch}{1.2}
\begin{tabular}{
l
rrr@{\hspace{5pt}}
rrr@{\hspace{5pt}}
rrr
}
\toprule
\multirow{2}{*}{Context}
& \multicolumn{3}{c}{GLM-5.2}
& \multicolumn{3}{c}{DeepSeek-V4-Flash}
& \multicolumn{3}{c}{Hy-4-Preview} \\
\cmidrule(lr){2-4}
\cmidrule(lr){5-7}
\cmidrule(lr){8-10}
& Native & \topk & Speedup
& Native & \topk & Speedup
& Native & \topk & Speedup \\
\midrule
256K & 4.17  & 4.07  & 1.02$\times$
     & 2.18  & 2.16  & 1.01$\times$
     & 5.62  & 5.44  & 1.03$\times$ \\
512K & 9.45  & 8.83  & 1.07$\times$
     & 5.13  & 4.81  & 1.07$\times$
     & 13.20 & 12.04 & 1.10$\times$ \\
768K & 16.30  & 14.62 & 1.11$\times$
     & 8.87  & 7.85  & 1.13$\times$
     & 24.57 & 19.81 & 1.24$\times$ \\
1M   & 24.62 & 21.11 & 1.17$\times$
     & 13.61 & 11.60 & 1.17$\times$
     & 41.71 & 28.02 & 1.49$\times$ \\
\bottomrule
\end{tabular}

\end{table*}

\begin{table}[t]
  \centering
  
  \caption{
    Joint exact-match accuracy (\%) on the two-needle retrieval test, averaged over all needle positions.
    The four context-length buckets contain 50, 20, 10, and 5 test
    instances, respectively.
  }
  
  \label{tab:two_needle}

  \small
  \setlength{\tabcolsep}{3pt}
  \renewcommand{\arraystretch}{1}

  \begin{tabular*}{0.9\columnwidth}{
    @{\extracolsep{\fill}}llrrrr@{}
  }
    \toprule
    Model & Method
    & \multicolumn{1}{c}{64K--128K}
    & \multicolumn{1}{c}{128K--256K}
    & \multicolumn{1}{c}{256K--512K}
    & \multicolumn{1}{c}{512K--1M} \\
    \midrule

    \multirow{2}{*}{GLM-5.2}
      & Native  & 96.10 & 90.86 & 90.25 & 63.16 \\
      & \topk & 96.11 & 90.89 & 90.25 & 63.12 \\

    \addlinespace[2pt]

    \multirow{2}{*}{DeepSeek-V4-Flash}
      & Native  & 98.10 & 95.37 & 91.44 & 86.51 \\
      & \topk & 98.10 & 95.37 & 91.44 & 86.50 \\
    \addlinespace[2pt]

    \multirow{2}{*}{Hy-4-Preview}
      & Native  & 94.42 & 81.98& 47.11 & 42.70 \\
      & \topk & 94.45 & 81.99& 47.40 & 42.70 \\
    \bottomrule
  \end{tabular*}
\end{table}

\noindent\textbf{End-to-End Long-Context Prefilling.} We integrate \topk into three widely used LLM systems: vLLM, SGLang, and Megatron-LM. In each system, we replace only the native Indexer-TopK kernel with \topk, while keeping all other framework components and settings unchanged within each paired comparison. We evaluate the FP8 checkpoints of three models with native support for contexts of up to 1M tokens: GLM-5.2~\cite{glm5team2026glm5vibecodingagentic}, DeepSeek-V4-Flash~\cite{deepseekai2026deepseekv4}, and Hy-4-Preview~\cite{tencenthyteam2026hy4preview}. We construct the input prompts from the Salesforce WikiText dataset and use a batch size of one.

Because context parallelism (CP) support in vLLM and Megatron-LM is incomplete, we evaluate all three frameworks with expert parallelism (EP) and tensor parallelism (TP). We separately evaluate SGLang with CP enabled to examine its effect on the speedup achieved by \topk. For a fair comparison, Each Native-\topk pair uses the same model-specific parallel configuration. The complete configurations are provided in Appendix Table~\ref{tab:parallel-config}.


\noindent\textbf{Exactness and Model Quality.}
We assess whether \topk preserves model-level quality and kernel-level exactness. At the model level, we use a two-needle retrieval test based on the needle-in-a-haystack protocol, which is widely adopted to evaluate the long-context capabilities of large language models~\citep{hsieh2024ruler}. The task inserts two target facts at controlled positions within a long distractor context and poses a question that can be answered correctly only by retrieving both facts. We use identical prompts and greedy-decoding settings for the native implementation and \topk, vary the context length up to one million tokens, and report joint exact-match accuracy and output-token agreement. At the kernel level, we compare the Top-$k$ indices  produced by \topk with those of the native implementation and report Top-$k$ recall.

\noindent\textbf{Sparse-Attention Kernel Benchmark.}
We next isolate the DSA workload to examine the source of the end-to-end improvements.
We compare \topk separately against two implementations from vLLM and SGLang. The benchmark inputs is constructed from the first-layer attention activations of GLM-5.2, using text sampled from Wikipedia~\citep{merity2016pointer}. We consider context lengths of 256K, 512K, 768K, and 1M tokens, vary the prefill chunk size from 128 to 8,192 tokens, and fix $k$ to 2,048. For each configuration, we report the aggregate kernel latency for processing 8,192  tokens. When the queries are divided into multiple chunks, the reported latency is the sum of the latency across all chunks. We additionally measure peak auxiliary memory consumption. Although peak memory overheads are impractical for production deployment, we retain these configurations to characterize the best latency achievable by the baselines when sufficient memory is available.

We separately benchmark \dsa{} against the native sparse-attention kernel used by vLLM on the same GLM-5.2 activation workload, with $k=2{,}048$ and context lengths of 256K, 512K, 768K, and 1M. Each attention-kernel measurement processes one 8,192-token query chunk; these latencies exclude Indexer-TopK scoring and selection.


\noindent\textbf{Large-Scale Retrieval Benchmark.}
Finally, we evaluate \topk on large-scale vector retrieval to determine whether its efficiency gains generalize beyond sparse attention. We construct a corpus of five million passages sampled from MSMARCO-V2.1, a widely used text-retrieval dataset. Each passage is encoded as a 768-dimensional vector using the Snowflake Arctic-embed-m-v1.5 model~\citep{snowflake2024embedding}.
We construct the query set $Q$ by randomly sampling 1,000 passages and evaluating corpus sizes of 1M, 2M, 4M, and 5M vectors. We vary $k$ over 128, 512, 1,024, 4,096, and 8,192 and use a batch size of 64 to emulate concurrent retrieval requests. 
We 
compare \topk with 
Torch on this task. 


\noindent\textbf{Implementation.} All our implementations are written in CUDA C++ using CuTe/CUTLASS, while the baselines are evaluated base on their official implementations. The experiments are conducted on an NVIDIA B200 GPU with 179 GB of memory using CUDA 12.8, while our method can also be extended to other GPU architectures. 
All reported running times are averaged over 20 runs.

\begin{figure*}[t]
\centering

\subcaptionbox{DSA Prefill Indexer-TopK Kernel\label{fig:dsa}}{

\includegraphics[width=0.9\textwidth]{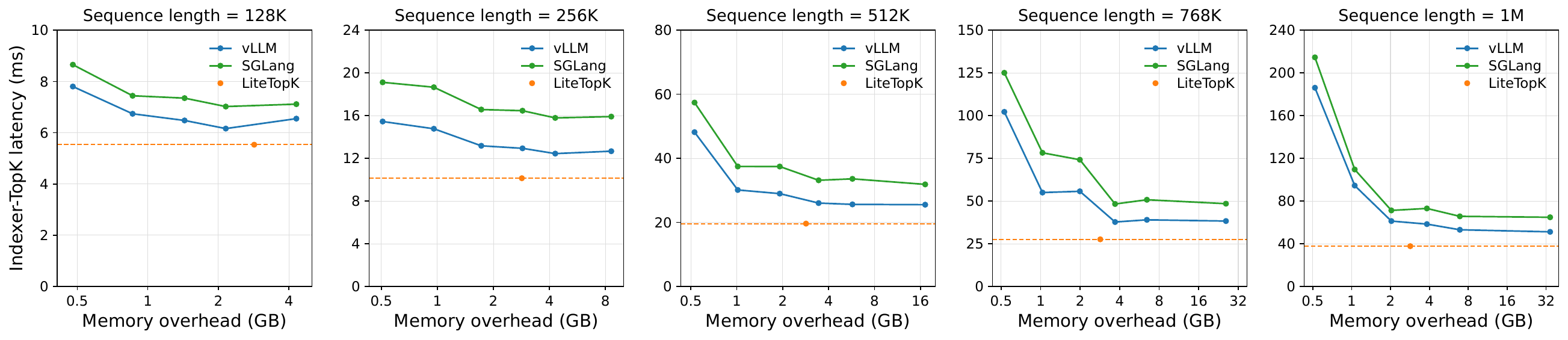}
}
\subcaptionbox{DSA Decode Indexer-TopK Kernel\label{fig:dsa_decode}}{

\includegraphics[width=0.9\textwidth]{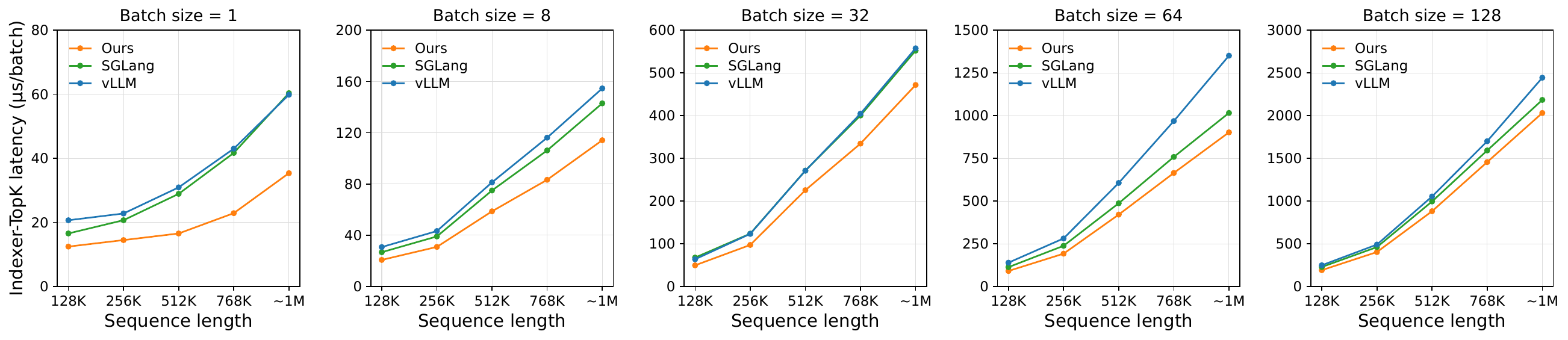}
}
\subcaptionbox{Vector Retrieval\label{fig:msmarco}}
{

\includegraphics[width=0.9\textwidth]{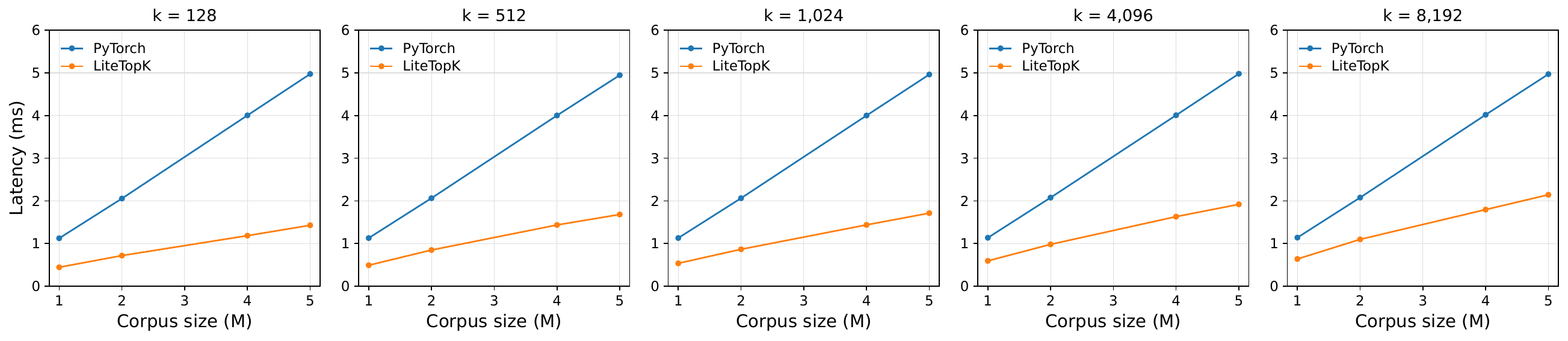}
}

\caption{Runtime comparison between \topk and existing methods on sparse-attention and retrieval workloads.}

\label{fig:kernel-runtime}
\end{figure*}

\subsection{Experimental Results}
\label{sec:exp-results}


\noindent\textbf{End-to-End Long-Context Prefilling.}
Table~\ref{tab:e2e_frameworks} reports end-to-end prefill latency on eight B200 GPUs. LiteTopK reduces latency across all evaluated models, systems, and context length configurations, with speedups increasing monotonically with context length. In particular, at a context length of 1M tokens, it achieves speedups of 1.21--1.37$\times$ for GLM 5.2, 1.10--1.38$\times$ for DeepSeek-V4-Flash, and 1.19--1.27$\times$ for Hy-4-Preview.  Table~\ref{tab:e2e_sglang_cp} further shows that \topk continues to provide end-to-end prefill speedups with context parallelism enabled. At a 1M-token context length, it achieves speedups of 1.17$\times$ for both GLM-5.2 and DeepSeek-V4-Flash, and 1.49$\times$ for Hy-4-Preview. The larger gain for Hy-4-Preview is attributable to its use of BF16 in attention for numerical stability and CP configuration, which leaves less HBM capacity available for the Indexer-TopK workspace and makes \topk's memory savings particularly beneficial.



\noindent\textbf{Exactness and Model Quality.} 
We first evaluate whether \topk preserves the outputs of the native Indexer-TopK 
operator. Across all evaluated configurations, \topk achieves 100\% recall relative to the native implementation. 
At the model level, Table~\ref{tab:two_needle} compares joint exact-match accuracy on the two-needle retrieval task. Across all three models and four context-length ranges, The maximum absolute difference is 0.29 percentage points. Small fluctuations also occur across repeated runs of the native vLLM and SGLang implementations, suggesting that the observed differences reflect run-to-run variability rather than a systematic degradation caused by \topk. Overall, \topk maintains nearly identical long-context retrieval accuracy.

\noindent\textbf{Sparse-Attention Indexer-TopK Kernel Performance.}
Figure~\ref{fig:dsa} compares the Indexer-Topk kernel latency of \topk, vLLM, and SGLang on a single NVIDIA B200 GPU.
With a fixed query batch of 8,192 tokens and a context length of 1M, \topk achieves a latency of 37.78\,ms, compared with 51.25\,ms for vLLM and 64.85\,ms for SGLang, corresponding to speedups of $1.36\times$ and $1.72\times$, respectively.
At 256K, the corresponding speedups are $1.25\times$ and $1.57\times$, respectively. 
These results show that \topk's relative advantage increases with context length in the evaluated configurations. 

\topk also substantially reduces memory overhead. At 1M, processing all 8,192 query tokens in a single subcall requires approximately 32\,GB of auxiliary workspace for the baselines, compared with 2.84\,GB for \topk. Reducing the baseline query chunk size lowers its memory footprint but increases the total latency. For example, vLLM takes 60.15\,ms and 54.69\,ms with query chunks of 1,024 and 1,536 tokens, using about 4\,GB and 7\,GB of workspace, respectively. \topk can process the full query batch 
with a smaller memory footprint. 

\begin{figure}[t]
\centering
\includegraphics[width=0.8\linewidth]{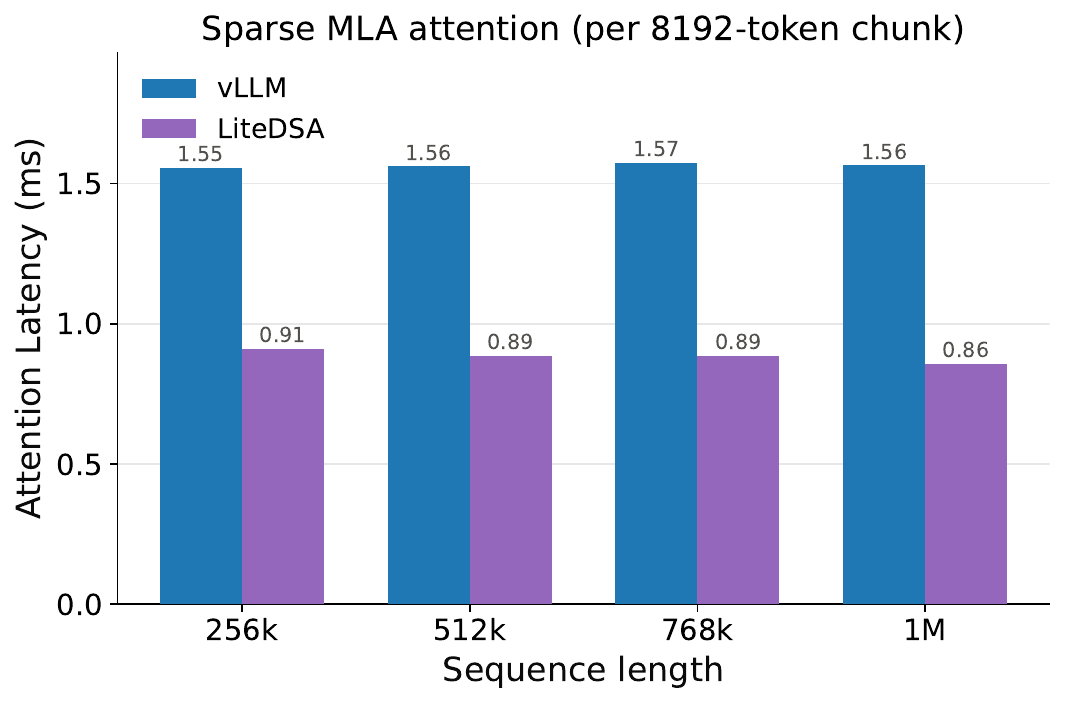}
\caption{Sparse-attention kernel latency of native vLLM and \dsa{} on an NVIDIA B200 GPU. Each measurement processes an 8,192-token query chunk with $k=2{,}048$ and excludes Indexer-TopK.}
\label{fig:kernel-att}
\end{figure}

\noindent\textbf{Sparse-Attention Kernel Performance of \dsa.}
Figure~\ref{fig:kernel-att} compares \dsa{} with the native vLLM sparse-attention kernel. Across context lengths of 256K, 512K, 768K, and 1M, \dsa{} reduces latency from 1.55, 1.56, 1.57, and 1.56\,ms to 0.91, 0.89, 0.89, and 0.86\,ms per 8,192-token query chunk, respectively. This corresponds to speedups of $1.70\times$--$1.81\times$. Attention latency varies little with the full context length because each query attends to its selected 2,048 candidates. Sharing KV loads across neighboring queries reduces redundant memory traffic while the per-query membership masks retain the original candidate sets.

\noindent\textbf{Decode Indexer-TopK Kernel Performance.} 
Figure~\ref{fig:dsa_decode} compares the decode Indexer-TopK latency of \topk, vLLM, and SGLang on a single NVIDIA B200 GPU. 
Across all evaluated configurations, \topk{} achieves the lowest latency, delivering speedups of $1.08\times$--$1.82\times$ over SGLang and $1.13\times$--$1.88\times$ over vLLM. At batch size 1, the speedup over SGLang increases from $1.33\times$ at a context length of 128K to $1.82\times$ at 768K. At batch size 128, the corresponding speedup decreases from $1.19\times$ at 128K to $1.08\times$ at approximately 1M. The smaller gains at larger batch sizes are expected: indexer score computation accounts for a growing share of total decode latency, reducing the fraction attributable to TopK selection and limiting the speedup achievable with \topk{}. Table~\ref{tab:e2e_decode} reports end-to-end decode speedups of up to $1.04\times$. These gains are modest because decoding is primarily memory-bound, and reducing Indexer–TopK overhead addresses only a small fraction of the overall memory-access cost.

\noindent\textbf{Large-Scale Retrieval Performance.}
Figure~\ref{fig:msmarco} compares \topk and PyTorch kernel latency for inner-product computation and top-$k$ selection on MS MARCO passage embeddings, covering 1M--5M vectors and $k \in \{128, 512, 1024, 4096, 8192\}$. \topk achieves $1.79\times$--$3.48\times$ speedups across all configurations.
In particular, at 5M vectors, it reduces latency from 4.97\,ms to 1.43\,ms for $k=128$, leading to 3.48$\times$ speedup.






\begin{table*}[htbp]
\centering
\small
\setlength{\tabcolsep}{4pt}

\caption{Decode latency (s) of GLM-5.2 with SGLang on eight B200 GPUs. 
}

\label{tab:e2e_decode}
\resizebox{0.95\textwidth}{!}{%
\begin{tabular}{@{}rrc rrc rrc rrc@{}}
\toprule
\multicolumn{3}{c}{256K} & \multicolumn{3}{c}{512K}
& \multicolumn{3}{c}{768K} & \multicolumn{3}{c}{1M} \\
\cmidrule(lr){1-3} \cmidrule(lr){4-6} \cmidrule(lr){7-9} \cmidrule(lr){10-12}
Native & \topk & Speedup & Native & \topk & Speedup
& Native & \topk & Speedup & Native & \topk & Speedup \\
\midrule
1.20 & 1.17 & 1.03$\times$ & 1.24 & 1.21 & 1.03$\times$
& 1.30 & 1.26 & 1.03$\times$ & 1.36 & 1.31 & 1.04$\times$ \\
\bottomrule
\end{tabular}}

\end{table*}

\IfFileExists{content/5.4-predictive-e2e.tex}{
}{}

\section{Conclusion and Discussion} This paper presents \topk{}, the first fused Indexer-TopK framework that exploits the concentration of sparse-attention scores to avoid materializing the full score matrix. We further introduce \dsa{}, which exploits overlap among neighboring queries' selected candidates to reuse key/value loads in downstream sparse attention, with per-query masks preserving the original candidate sets. 


\bibliography{ref}
\bibliographystyle{iclr2027_conference}

\newpage
\appendix
\section*{Appendix}

\section{DSA Score Concentration Across Models and Layers}
\label{sec:pattern_appendix}
\begin{figure}[htbp]
  \centering
  \includegraphics[width=\linewidth]{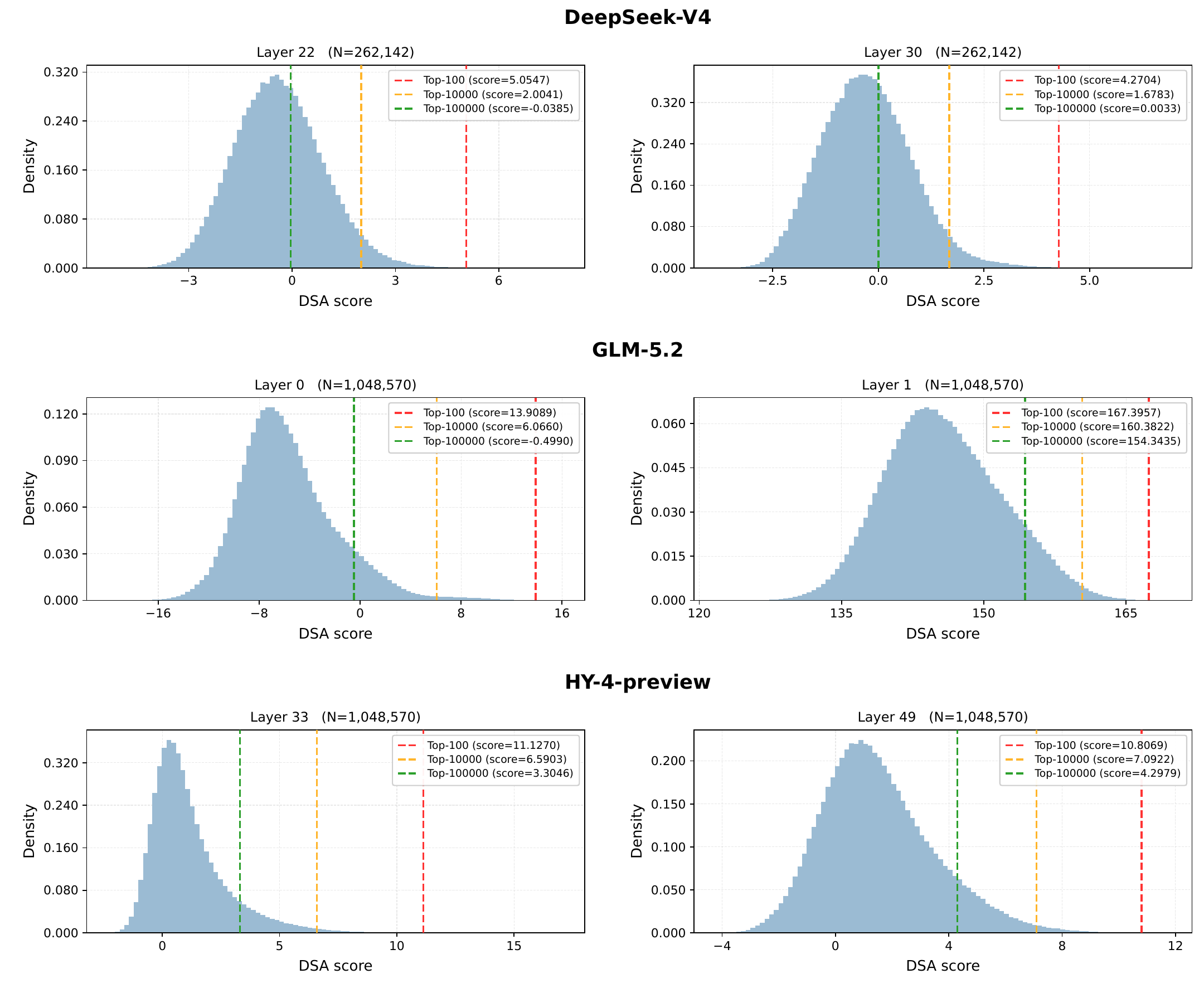}
  \caption{
    \textbf{DSA score distributions across models and layers.}
    The three rows correspond to DeepSeek-V4 (layers 2 and 42),
    GLM-5.2 (layers 0 and 14), and HY-4-preview (layers 0 and 77).
  }
  \label{fig:dsa-score-concentration}
\end{figure}

Figure~\ref{fig:dsa-score-concentration} compares DSA score distributions from two selected layers of each of three models: DeepSeek-V4, GLM-5.2,
and HY-4-preview. All six model--layer pairs exhibit a similar
score-concentration pattern that \topk{} can exploit. Negative scores arise because ReLU is applied to the per-head dot products,
while the head weights can be negative, making their weighted sum potentially negative.

\section{Model-specific configurations}
Appendix Table~\ref{tab:parallel-config} summarizes the model-specific and framework-specific parallel configurations used in our end-to-end evaluation, which are held fixed within each paired Native--\topk comparison.

\begin{table}[t]
    \centering
    \caption{
        Model-specific parallel configurations used in the end-to-end
        experiments. Each paired Native--\topk comparison uses an identical
        configuration. A dash indicates that no qualified result was available.
    }
    \label{tab:parallel-config}
    \small
    \setlength{\tabcolsep}{10pt}
    \begin{tabular}{@{}lccc@{}}
        \toprule
        Model & vLLM & SGLang & Megatron-LM \\
        \midrule
        GLM-5.2
        & TP8 / EP8 
        & TP8 / EP8
        & TP8 / EP8 \\
        
        DeepSeek-V4-Flash
        & TP8 / EP1
        & TP8 / EP1
        & TP8 / EP1 \\
        
        Hy-4-Preview
        & TP8 / EP8
        & TP8 / EP8
        & -- \\
        \bottomrule
    \end{tabular}
\end{table}

\end{document}